# Curriculum-style Local-to-global Adaptation for Cross-domain Remote Sensing Image Segmentation

Bo Zhang, Tao Chen, *Member, IEEE*, and Bin Wang, *Senior Member, IEEE*

*Abstract*—Although domain adaptation has been extensively studied in natural image-based segmentation task, the research on cross-domain segmentation for very high resolution (VHR) remote sensing images (RSIs) still remains underexplored. The VHR RSIs-based cross-domain segmentation mainly faces two critical challenges: 1) Large area land covers with many diverse object categories bring severe local patch-level data distribution deviations, thus yielding different adaptation difficulties for different local patches; 2) Different VHR sensor types or dynamically changing modes cause the VHR images to go through intensive data distribution differences even for the same geographical location, resulting in different global feature-level domain gap. To address these challenges, we propose a curriculum-style local-to-global cross-domain adaptation framework for the segmentation of VHR RSIs. The proposed curriculum-style adaptation performs the adaptation process in an easy-to-hard way according to the adaptation difficulties that can be obtained using an entropy-based score for each patch of the target domain, and thus well aligns the local patches in a domain image. The proposed local-to-global adaptation performs the feature alignment process from the locally semantic to globally structural feature discrepancies, and consists of a semantic-level domain classifier and an entropy-level domain classifier that can reduce the above cross-domain feature discrepancies. Extensive experiments have been conducted in various cross-domain scenarios, including geographic location variations and imaging mode variations, and the experimental results demonstrate that the proposed method can significantly boost the domain adaptability of segmentation networks for VHR RSIs. Our code is available at: https://github.com/BOBrown/CCDA_LGFA.

## I. INTRODUCTION

WITH the rapid development of Earth observation technologies, a large number of very high resolution (VHR) remote sensing images (RSIs) are becoming widely available to monitor the changing status of a large land surface [1], [2]. Semantic segmentation for VHR RSIs aims to assign a category label to every pixel in the VHR RSIs, which is a crucial step for the understanding of ground information and plays an important role in many urban-oriented applications such as traffic management, land surveying, city planning, and environmental monitoring [3]-[8].

Recently, benefitting from the constant progress of convolution neural network (CNNs) [9], [10], the ability to perform the pixel-wise image segmentation from given input images has been pushed forward a lot [11]-[15], on condition that training and test samples are strictly drawn from the same data distribution. However, for semantic segmentation of VHR RSIs, this assumption is difficult to be guaranteed, largely due to that the data distribution of VHR RSIs tends to be different since they are acquired from different imaging sensors and geographical locations. Besides, semantic segmentation of VHR RSIs is required to work reliably and accurately across different sensors and other urban scenes. Unfortunately, the generalization ability of data-driven segmentation networks [11]-[16] well-trained on their original source domain would be eventually degenerated when deployed to a new target domain with data distribution differences (domain shifts).

Retraining the segmentation network on the new target domain with data distribution differences may improve the segmentation accuracy and achieve satisfactory results on the new domain. However, data collected from the target domain still needs to be manually annotated [17]. Especially for VHR RSIs, performing such a pixel-level manual annotation process is highly dependent on knowledge from professionals in the field of remote sensing, which makes the manual annotation process more time-consuming [18]. Therefore, it is necessary to transfer the VHR RSIs-based segmentation network from a well-labeled source domain to an unlabeled target domain.

One common solution to alleviate the data discrepancy is unsupervised domain adaptation (UDA) [19]-[26], whose goal is to reduce the inter-domain feature discrepancy in an unsupervised way that only utilizes labeled source samples and unlabeled target samples. Most existing works can be roughly categorized into two classes: metric discrepancy-based methods and adversarial learning-based methods. The metric discrepancy-based methods minimize the feature distribution differences between domains with some metric criteria including maximum mean discrepancy (MMD) [19], [20], deep correlation alignment (CORAL) [21], and multi-kernel MMD (MK-MMD) [22]-[23], while the adversarial learning-based methods try to learn domain-invariant features by means of a domain discriminator [24]-[30]. However, it should be noted that the above methods [19]-[26] focus on the cross-domain adaptation in image classification task, where only a single

Manuscript received July 15, 2021; revised August 26, 2021; accepted September 30, 2021. This work was supported by the National Natural Science Foundation of China under Grant 61971141 and Grant 61731021. *(Corresponding author: Bin Wang)*

The authors are with the Key Laboratory for Information Science of Electromagnetic Waves (MoE), Fudan University, Shanghai 200433, China, and also with the Research Center of Smart Networks and Systems, School of Information Science and Technology, Fudan University, Shanghai 200433, China (e-mail: bo.zhangzx@gmail.com).

Digital Object Identifier 10.1109/TGRS.2021.3117851

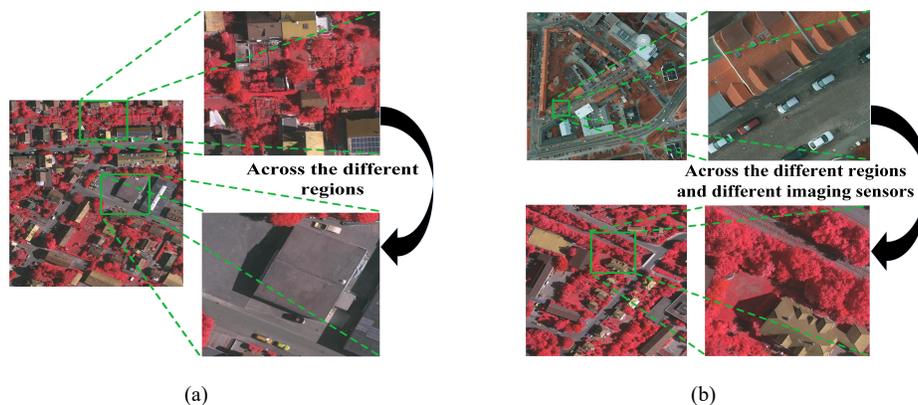

(a) (b)

Fig.1. Illustration of distinct characteristics of data in VHR RSIs: (a) Challenge one: the two patches (512×512 pixels) sampled from the same VHR RSI of Vaihingen IR-R-G dataset (denoting the target domain) have considerable data discrepancies; (b) Challenge two: the two patches sampled from different imaging sensors on different geographic locations also present domain discrepancies.

object of interest exists in the whole input image. As a result, these classification-based works fail to reduce the inter-domain discrepancies of local detail features, and ignore the importance of spatially structural relations among multiple semantic categories.

Driven by the success of UDA methods in the image classification task, some researchers start to extend the study of cross-domain adaptation from the basic image classification task to the high-level downstream tasks such as object detection [31], [32] and semantic segmentation [33]-[39], *via* pixel-level image-to-image translation ways [40]-[44] or feature-level adversarial learning-based adaptation ways [31]-[37]. For example, in order to obtain high cross-domain segmentation accuracy, an easy task that aims to predict the global label distributions is first performed, and further, the network training of the segmentation model is effectively regularized [33], [34]. AdaptSegNet [35] focuses on aligning the structured output space during the adaptation process and ProDA [37] tackles the problem of UDA from the perspective of self-training by using the feature centroids to learn the compact target structure. However, these methods are difficult to address the domain discrepancies residing in VHR RSIs. This is mainly due to that VHR RSIs between different domains usually have their own distinct characteristics such as the imaging mode (sensor) differences, color saturation variations, and spatial layout changes.

For the above reasons, in the field of remote sensing image semantic segmentation, a few attempts [45]-[50] have been made to design a cross-domain adaptation framework under dynamically changing imaging modes or land-cover categories. For example, by matching the cross-domain features from a two-branch structure consisting of a lightweight UNet and a discrete/inverse wavelet network to adapt the segmentation model, BSANet [45] can alleviate the domain discrepancies of feature distributions and further boost the segmentation accuracy of VHR RSIs. Besides, a full-space domain adaptation framework is developed to match the inter-domain discrepancies using adversarial learning from the image space, feature space, and output space [48]. Recently, DualGAN [49] is proposed to generate some target-domain-like RSIs, by imposing multiple weakly-supervised constraints on the final image-to-image style translation process to train a domain adaptive segmentation network. Although these works [45]-[50] are inspiring and have achieved performance gains on the target domain, they still face two important challenges as follows.

First, since VHR RSIs acquired from the same imaging sensor often contain a wide range of land covers, the patches sampled from such RSIs usually present distinct intra-domain distribution differences and different cross-domain adaptation difficulties of local patches within a domain image. For example, Fig. 1(a) shows such a considerable patch-wise domain discrepancy on the target image, where the two patches that are sampled from different geographical locations of the same VHR RSI have significant feature differences. As a result, the cross-domain adaptation process should consider the intrinsic feature differences within a domain (or a VHR RSI).

Second, the VHR RSIs collected from different imaging sensors on different geospatial regions often present domain discrepancies in both local detailed features (such as local appearance, texture, and semantic changes of land-cover semantic categories) and global structural features (such as spatial layout discrepancies of land-cover semantic categories). The above two domain discrepancies are well demonstrated in Fig. 1(b), where the appearance/texture cross-domain differences in "Tree" can be obviously observed, and the geometric layout of "Building" (or structural distribution of land-cover types) is dynamically changing between domains. For these reasons, learning a domain-invariant feature extractor that can ensure both good local detailed and global structural consistency of source-target feature representations is indispensable for VHR RSIs-based semantic segmentation.

To overcome the above problems, in this paper, we propose a curriculum-style local-to-global cross-domain semantic segmentation framework for VHR RSIs. Specifically, considering that the patches sampled from the same target image often have diverse data variations within the domain, we propose that the cross-domain adaptation for all the target patches should follow the idea that easy-to-adapt patches need to be adapted before the difficult ones during the adaptation process, and develop a curriculum-style cross-domain adaptation (CCDA) strategy. By calculating entropy-based scores to sort the target patches according to their degree of difficulty, some easy-to-adapt target patches can be selected to perform the initial source-target domain adaptation. Next, the initially aligned segmentation model can be further adapted to hard-to-adapt patches based on both the pseudo-labeled easy-to-adapt target patches and labeled source patches. Furthermore, during the feature learning process of the segmentation model, to reduce the cross-domain feature differences in both local details and spatial layouts of VHR RSIs, a local-to-global feature alignment (LGFA) module is designed. We consider that the feature difference alignment should also follow the similar easy-to-hard adaptation way, where the global spatial layout of all semantic categories is often more complicated compared with local semantic information that describes relatively local patterns of semantic objects [35], [36]. Therefore, the domain discrepancies of features representing local semantic details of an object should be firstly aligned using adversarial learning between the segmentation backbone and the developed domain discriminator. Next, based on the well-aligned features in local semantic information, we further align the source-target entropy maps representing class-specific predictions which can describe global structural information of land-cover semantic categories.

We conduct extensive experiments on several common benchmarks [51] under two typical cross-domain scenarios: 1) domain adaptation under different geographic locations and 2) domain adaptation under both different geographic locations and imaging modes. All experimental results demonstrate the effectiveness of the proposed curriculum-style adaptation strategy in alleviating the domain discrepancies of VHR RSIs, and it can also significantly boost the domain adaptability for RSIs-based segmentation task. Further, we conduct experiments employing another baseline model to validate the generalization capability of the proposed method under different baseline segmentation models.

The main contributions of this paper can be concisely summarized as follows:
1) We reveal a crucial aspect to successfully adapt RSIs-based segmentation network from its source domain to target domain, namely, curriculum-style adaptation strategy that aims to progressively adapt the source network according to the uncertainty of target patches to be adapted. This should be a pioneer work to consider the domain adaptation from the perspective of curriculum learning for the segmentation of VHR RSIs.
2) Aiming at alleviating the domain discrepancies of RSIs in both local semantic and global spatial layout information, we design semantic-level and entropy-level domain classifiers to respectively quantify the two discrepancies so that both the locally semantic and globally structural feature differences between domains can be fully reduced.

## II. The Proposed Method

The purpose of this work is to transfer a pre-trained segmentation network on the RSIs-based source domain to a RSIs-based unlabeled target domain in some typical cross-domain scenarios including different imaging modes and geographic locations. The overall framework of the proposed method is shown in Fig. 2, which performs the source-target domain adaptation through the following CCDA strategy. First, the given source-domain model is adapted from the source domain to the selected easy-to-adapt patches of the target domain. Next, initialized from the model that is trained via the stage one, the pseudo-labels of these easy-to-adapt target patches can be obtained, and the stage two can be performed in order to further adapt the model to hard-to-adapt target patches. During each stage above, the LGFA module (consisting of a semantic-level domain classifier and an entropy-level domain classifier) is plugged into the baseline segmentation network to achieve the locally semantic and globally structural feature matching.

To better illustrate the curriculum-style domain adaptation process for RSIs segmentation, we first give the problem definition of the cross-domain adaptation and semantic segmentation. Next, we give detailed descriptions of the proposed curriculum-style local-to-global adaptation. Then, the detailed network structures of designed semantic-level and entropy-level domain classifiers are illustrated. Finally, we give the overall optimization objective and adaptation strategy of the proposed method.

### A. Preliminaries

*1) Unsupervised Domain Adaptation:* Suppose that $x$ is an input image, $I$ is its representation where $I = B(x)$ and $B$ is a backbone of a segmentation network, *e.g.*, ResNet-101 in Deeplab-v3 [14], and $Y$ denotes the pixel-level annotation in semantic segmentation task, respectively. The purpose of unsupervised domain adaptation is to learn a generalized $B$ between domains, so that the $B$ can be safely adapted to a new target domain $t$, where the labeled images from the source domain are given while the annotation $Y_t$ of images from the target domain is unavailable.

*2) Source-domain Semantic Segmentation:* Typical CNNs based semantic segmentation model is generally composed of a backbone network $B$ that transforms the input image from the pixel space into high-level representations and a pixel-level classification module $P$ that aims to convert the above representations into a class-specific output space. Given the labeled image $x_s$ from the source domain $s$ and its pixel-level annotation $Y_s$, the objective of semantic segmentation is to optimize the $B$ and $P$ so that the segmentation loss is minimized on the source domain data. This process can be formulated as follows:

$$L_{seg}(B,P) = -\sum_{h=1}^{H}\sum_{w=1}^{W}\sum_{c=1}^{C} Y_s^{(h,w,c)} \log \sigma(P(B(x_s)))^{(h,w,c)} \quad (1)$$

where $P(B(x_s)) \in \mathbb{R}^{H \times W \times C}$ denotes the class-specific output over $C$ classes, and $H$ and $W$ represent the height and width of the output, respectively, and $\sigma$ denotes the Softmax activation function [12], [13], which is commonly used in semantic segmentation.

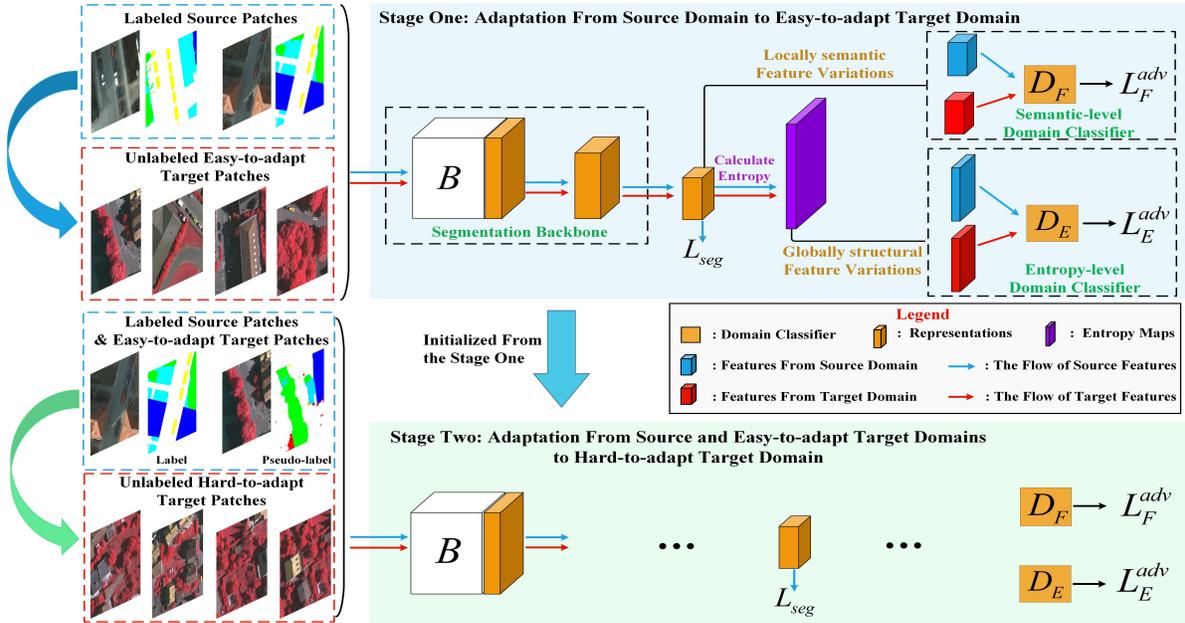

Fig. 2. The overview of the proposed method.

### B. Curriculum-style Local-to-global Cross-domain Adaptation

To better illustrate the curriculum-style local-to-global cross-domain adaptation, we first introduce an entropy-based ranking method to split all the target patches into easy or hard patches. Then, the first and second stages of cross-domain adaptation are described in detail.

*1) Entropy-based Patch-wise Ranking:* The purpose of the proposed method is to progressively adapt all the patches from the target domain image according to their degree of difficulty. However, the annotation from the target domain is not allowed to be obtained during the whole adaptation process. Thus, we have to employ the entropy map of the segmentation prediction to represent the uncertainty of every patch from the target domain $t$, which can be regarded as an unsupervised manner. Given the CNNs based segmentation model with $B$ and $P$ well-trained on the source domain $s$, the entropy map $M$ of each target patch can be calculated as follows:

$$M(P(B(x_t))) = -\sigma(P(B(x_t))) \log \sigma(P(B(x_t))) \quad (2)$$

where $\sigma$ represents the Softmax activation function [12], [13], which can convert the high-level semantic features into the prediction probability over $C$ classes.

Further, the cross-domain uncertainty score assigned to each target patch can be obtained by calculating the mean value $m$ of the entropy map $M$ as follows:

$$m = \frac{1}{HWC} \sum_{h=1}^{H} \sum_{w=1}^{W} \sum_{c=1}^{C} M^{(h,w,c)}. \quad (3)$$

According to the calculated patch-wise uncertainty score $m$, we can select some easy-to-adapt target patches to perform the first stage of the adaptation process, given a preset proportion of easy-to-adapt patches in the total target patches $\gamma$.

*2) Cross-domain Adaptation During the First Stage:* Due to that the patches from the same remote sensing image of the target domain present a certain difference in data distribution, the cross-domain adaptation needs to consider such a difference. For this reason, we perform a two-stage cross-domain adaptation to gradually boost the segmentation accuracy on the target domain. For the first stage of cross-domain adaptation, we try to adapt the source-domain segmentation network from the original source patches to some easily adapted target patches.

The domain discrepancies of RSIs mainly arise from two aspects: locally semantic feature variations such as cross-domain differences in local texture/appearance, and globally spatial layout changes such as cross-domain differences in global spatial structure. Thus, to successfully adapt the RSIs-based segmentation network under such domain discrepancies, we need to consider both local and global variations of semantic features and spatial information, respectively. To this end, we first design the adversarial-based semantic-level domain classifier to measure the cross-domain feature differences in semantics of local regions, and optimize it to reduce the domain gap as follows:

$$L_F^{adv}(D_F) = -\frac{1}{HW} \sum_h \sum_w \Big[ d \log \big( D_F(B(x_s)^{(h,w)}) \big) \\ + (1-d) \log \big( 1 - D_F(B(x_t)^{(h,w)}) \big) \Big] \quad (4)$$

where $D_F$ denotes the semantic-level domain classifier that is easily inserted into a commonly-used segmentation network with the domain label $d=1$ and $d=0$ for the source patches and the target patches, respectively, and $B(x_s)^{(h,w)}$ are the features located at $(h, w)$ of the feature map $B(x_s)$. The function of the domain classifier is to distinguish features from the source or target domains. Note that since the semantic-level domain classifier $D_F$ is designed in order to align semantic features located in a local region of an input patch, we employ a spatially-dense domain classifier to perform feature adaptation where the spatial resolution of features to be adapted can be preserved. Essentially it can encourage the $D_F$ to focus on learning the semantic differences in local regions of the given feature map $B(x)$, so as to alleviate the domain discrepancies caused by locally semantic feature variations. The visual network structure of the $D_F$ is illustrated in Fig. 3(a).

On the other hand, the cross-domain differences caused by the changes in global layout information related to multiple semantic categories can also degrade the segmentation accuracy on the target domain. For this reason, a RSIs-based cross-domain adaption framework should be able to address such feature variations. Since the **entropy map** calculated by the high-level semantic representations looks like edge detection results and has encoded the cross-domain spatial information of the corresponding semantic categories [36], we further develop an entropy-level domain classifier to measure the cross-domain data differences in globally spatial layouts as follows:

$$L_E^{adv}(D_E) = -\frac{1}{HW}\sum_h\sum_w \Big[ d \log\big(D_E(M(P(B(x_s))))^{(h,w)}\big) \quad (5)$$
$$+(1-d)\log\big(1-D_E(M(P(B(x_t))))^{(h,w)}\big)\Big]$$

where $D_E$ denotes the entropy-level domain classifier with the domain label $d=1$ and $d=0$ for the source and target patches, respectively, and $M(\cdot)$ represents the operation of calculating the entropy map. With the aid of adversarial learning, the globally spatial layout variations between different domains can be further reduced when the learned backbone network $B$ and pixel-level classification module $P$ successfully fool the entropy-level domain classifier $D_E$. Fig. 3(b) shows the detailed network structure of the $D_E$.

*3) Cross-domain Adaptation During the Second Stage:* During this stage, we further adapt the initially aligned segmentation model obtained by the first stage to the hard-to-adapt target patches.

As analyzed before, during this adaptation stage, both local and global cross-domain feature variations of the hard-to-adapt patches also should be considered, and they need to be fully aligned. Therefore, the proposed LGFA module is also inserted into the initially aligned segmentation model to alleviate the local-to-global feature differences residing in the hard-to-adapt patches.

Actually, due to that **the first stage** performs the adaptation from the source patches $x_s$ to easy-to-adapt target patches $x_t^{easy}$, the segmentation model which is trained via the first stage can be used to predict the pseudo-labels of these easy-to-adapt target patches $\hat{Y}_t^{easy}$. Further, driven by the pseudo-labeled easy-to-adapt target data $\{x_t^{easy}, \hat{Y}_t^{easy}\}$ that represent the approximate data distribution of the target domain and the labeled source data $\{x_s, Y_s\}$ that represent the actual data distribution of the source domain, the transferability of the remaining hard-to-adapt target patches $x_t^{hard}$ in RSIs is able to be further boosted, by means of adversarial learning between the learned segmentation network $\{B, P\}$ and the domain classifier $\{D_F, D_E\}$.

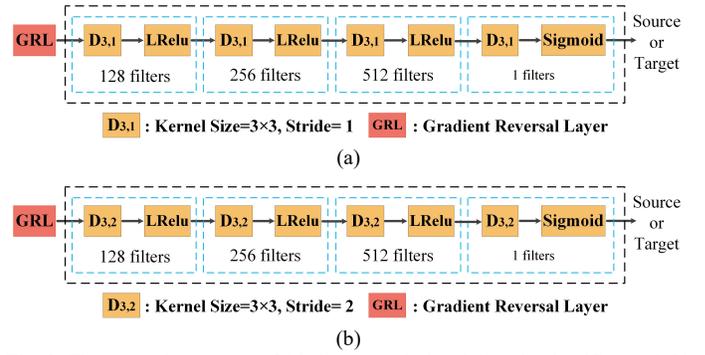

Fig. 3. The network structure of (a) the semantic-level domain classifier and (b) the entropy-level domain classifier. Gradient reversal layer (GRL) [24] is employed to simply implement the adversarial training process between the backbone and the domain classifier.

*C. Visual Network Structure of Multiple Domain Classifiers*

Both the semantic-level domain classifier $D_F$ and the entropy-level domain classifier $D_E$ use a four-layer network consisting of a convolution operation followed by the leaky ReLU (LReLU) activation function [14] for each layer. Note that we insert a Sigmoid activation function at the end of $D_F$ and $D_E$ since the purpose of the domain classifier is to distinguish features from the source domain or target domain. The $D_F$ employs a convolution operation with a kernel size of $3\times 3$ and a stride of $1$ in order to preserve the spatial resolution of the input features, while the stride of the convolution in the $D_E$ is set to $2$ in order to extract the global domain-specific representations. The visual network structure of our designed domain classifier is illustrated in Fig 3.

*D. Overall Loss Function and Curriculum-style Adaptation Strategy*

For the cross-domain semantic segmentation of VHR RSIs, the curriculum-style source-target adaptation of both patch level (denoting the adaptation from easy to hard patches) and feature level (denoting the adaptation from local to global feature variations) is necessary. Firstly, inter-domain discrepancies can be safely alleviated in the first stage of cross-domain adaptation by aligning the differences between the source patches and the easy-to-transfer patches, where cross-domain alignment of the hard-to-transfer patches will be ignored when the source-domain segmentation model is completely incapable to address the complicated cross-domain data discrepancies. This can actually avoid the so-called *negative transfer* phenomenon, which refers to that some non-transferable target patches are over-adapted. Secondly, the feature matching achieved by the proposed LGFA module also first aligns the local semantic feature differences between domains. Based on a well-aligned model in local semantic feature variations, the global feature differences will become relatively easier to be aligned.

*1) Optimization Objectives:* The overall loss function of the proposed method can be written as follows:

$$\min_{B,P} L_{seg}(B,P) \qquad (6)$$

$$\max_{D_F} \min_{\{B,P\}} L_{seg}(B,P) - \lambda L_F^{adv}(D_F) \qquad (7)$$

$$\max_{D_E} \min_{\{B,P\}} L_{seg}(B,P) - \lambda L_E^{adv}(D_E) \qquad (8)$$

where $\lambda$ denotes a hyper-parameter to trade off the importance between the model discriminability of the source patches $x_s$ and the model transferability of the target patches $x_t$.

*2) Curriculum-style Adaptation Strategy:* First, in order to encourage the segmentation model to learn segmentation related knowledge for subsequent transfer, the baseline segmentation model is trained only using the labeled image $x_s$ and its pixel-level annotation $Y_s$ from the source domain. Next, the curriculum-style adaptation is performed *in turn* according to the following stages.

**Adapt to Easy Target Patches Using $D_F$ and $D_E$:** The labeled source data $\{x_s, Y_s\}$ and unlabeled easy-to-adapt target patches $x_t^{easy}$ are mixed together for cross-domain alignment. First, jointly optimizing Eqs. (6) and (7) can ensure that the locally semantic feature discrepancies between domains are alleviated. Next, the model is further adapted by jointly optimizing Eqs. (6) and (8) to ensure the globally structural long-range feature consistency.

**Adapt to Hard Target Patches Using $D_F$ and $D_E$:** The locally semantic feature alignment for hard target patches can be performed using Eqs. (6) and (7) on the mixture set consisting of the labeled source data $\{x_s, Y_s\}$, pseudo-labeled easy-to-adapt target data $\{x_t^{easy}, \hat{Y}_t^{easy}\}$, and unlabeled hard-to-adapt target patch $x_t^{hard}$. Next, the globally structural feature differences are also further reduced by Eqs. (6) and (8).

## III. EXPERIMENTAL RESULTS

In this section, we first describe the datasets used to evaluate the proposed method. Then, we give the experimental setup for every adaptation stage, and introduce the evaluation metric. Further, we evaluate the proposed method in various domain adaptation scenarios including cross-domain semantic segmentation for different geographical locations and imaging modes.

*A. Dataset Description*

To evaluate the domain adaptability of the proposed method, we employ the two commonly-used aerial datasets including Potsdam and Vaihingen datasets, which are constructed by International Society for Photogrammetry and Remote Sensing (ISPRS) [51] and used as an important evaluation benchmark by a large number of RSIs-based semantic segmentation task [3]-[8], [45]-[50].

*1) Potsdam:* This dataset mainly covers a wide range of real-world land-cover types containing six semantic categories: clutter/background, impervious surfaces, car, tree, low vegetation, and building. The Potsdam dataset has different imaging modes including IR-R-G channels, R-G-B channels, and R-G-B-IR channels. To follow the experimental setting in [49], we use the first two imaging modes (IR-R-G/R-G-B) for the following experiments. Besides, The Potsdam dataset includes 38 very high resolution (VHR) True Orthophotos (TOP) images with 6000 × 6000 pixels.

*2) Vaihingen:* This dataset describes the same six semantic categories located in the Vaihingen city, containing only one imaging mode: IR-R-G channels. Besides, The Vaihingen dataset consists of 33 VHR TOP images with 2000 × 2000 pixels.

To train the segmentation network, we crop each image from the Potsdam or Vaihingen dataset into many patches with 512×512 pixels, generating 4598 and 1696 patches for Potsdam and Vaihingen dataset, respectively, which is consistent with the experimental setting in [45]-[50].

*B. Design of Baseline Models*

To fully verify the effectiveness of the proposed method, we first employ Deeplab-v3 [14] as the baseline segmentation model. Next, we devote to studying the impact of curriculum-style local-to-global cross-domain adaptation strategy on common segmentation networks, and design several models including Baseline with $D_F$ (denoting the original baseline model with the semantic-level domain classifier $D_F$) and SL-Adapted-Baseline with $D_E$ (denoting the semantically fully-aligned baseline model with the entropy-level domain classifier $D_E$).

*C. Experimental Setup and Evaluation Metric*

*1) Experimental Setup:* In order to learn sufficient source-domain knowledge for more effective cross-domain segmentation, we first train the baseline model on the labeled source domain. The baseline model is trained using stochastic gradient descent (SGD) optimizer with a batch size of 4, a momentum of 0.9, and a weight decay of 0.0005. The initial learning rate is set to 0.0005 and decays according to the polynomial annealing procedure proposed in [13].

Then, based on the baseline model which is well-trained on the source domain, we follow a curriculum-style local-to-global strategy to perform the cross-domain adaptation. For the two stages of cross-domain adaptation, the batch size for both the source and target domain data is set to 4, and the initial learning rate is set to 0.0025 and also decays according to the polynomial annealing procedure. Besides, the proportion of easy-to-adapt patches in the total target patches $\gamma$ is set to 0.5, and the hyper-parameter $\lambda$ is set to 0.1.

TABLE I
ADAPTATION RESULTS FROM POTSDAM IR-R-G TO VAIHINGEN IR-R-G. THE $D_F$ AND $D_E$ DENOTE THE SEMANTIC-LEVEL DOMAIN ADAPTATION AND ENTROPY-LEVEL DOMAIN ADAPTATION, RESPECTIVELY.

| Method | Clutter/background | | Impervious surfaces | | Car | | Tree | | Low vegetation | | Building | | Overall | |
|---|---|---|---|---|---|---|---|---|---|---|---|---|---|---|
| | IoU | F1-Score | IoU | F1-score | IoU | F1-score | IoU | F1-score | IoU | F1-Score | IoU | F1-score | mIoU | F1-score |
| Baseline (Deeplab-v3) [14] | 5.71 | 10.79 | 35.84 | 52.73 | 20.27 | 33.70 | 54.95 | 70.92 | 17.88 | 30.26 | 51.59 | 68.06 | 31.04 | 44.40 |
| BiSeNet only [11] | 1.67 | 3.28 | 38.46 | 55.55 | 9.42 | 17.20 | 45.54 | 62.57 | 23.04 | 37.44 | 28.15 | 43.93 | 24.38 | 36.65 |
| SEANet [39] | 11.11 | 20.00 | 45.57 | 62.59 | 32.71 | 49.29 | 50.42 | 66.83 | 23.06 | 37.45 | 57.99 | 73.27 | 36.81 | 51.60 |
| AdaptSegNet [35] | 4.60 | 8.76 | 54.39 | 70.39 | 6.40 | 11.99 | 52.65 | 68.96 | 28.98 | 44.91 | 63.14 | 77.40 | 35.02 | 47.05 |
| GAN-RSDA [50] | 2.12 | 4.15 | 39.88 | 57.02 | 8.20 | 15.15 | 26.56 | 41.97 | 26.53 | 41.94 | 40.97 | 58.10 | 24.04 | 36.40 |
| ProDA [37] | 3.99 | 8.21 | 62.51 | 76.85 | 39.20 | 56.52 | 56.26 | 72.09 | 34.49 | 51.65 | 71.61 | 82.95 | 44.68 | 58.05 |
| DualGAN [49] | **29.66** | **45.65** | 49.41 | 66.13 | 34.34 | 51.09 | **57.66** | **73.14** | 38.87 | 55.97 | 62.30 | 76.77 | 45.38 | 61.43 |
| Adaptation from Labeled Source Patches (LSPs) to Easy-to-adapt Target Patches (ETPs) | | | | | | | | | | | | | | |
| Baseline with $D_F$ (ours) | 4.79 | 9.34 | 61.70 | 76.59 | 39.06 | 56.39 | 56.04 | 72.48 | 37.12 | 54.51 | 70.06 | 82.11 | 44.80 | 58.57 |
| SL-Adapted-Baseline with $D_E$ (ours) | 4.50 | 8.67 | 66.09 | 79.05 | 39.75 | 56.94 | 56.55 | 73.00 | 41.53 | 58.63 | **76.90** | **86.69** | 47.55 | 60.49 |
| Adaptation from LSPs and Pseudo-labeled ETPs to Hard Target Patches | | | | | | | | | | | | | | |
| Baseline with $D_F$ (ours) | 13.66 | 25.63 | 67.54 | 79.93 | **44.99** | **62.21** | 52.40 | 69.65 | 45.41 | 62.69 | 76.21 | 86.32 | 50.03 | 64.41 |
| SL-Adapted-Baseline with $D_E$ (ours) | 20.71 | 31.34 | **67.74** | **80.13** | 44.90 | 61.94 | 55.03 | 71.90 | **47.02** | **64.16** | 76.75 | 86.65 | **52.03** | **66.02** |

TABLE II
ADAPTATION RESULTS FROM VAIHINGEN IR-R-G TO POTSDAM IR-R-G. THE DEFINITION OF $D_F$, $D_E$ FOLLOWS TABLE I

| Method | Clutter/background | | Impervious surfaces | | Car | | Tree | | Low vegetation | | Building | | Overall | |
|---|---|---|---|---|---|---|---|---|---|---|---|---|---|---|
| | IoU | F1-Score | IoU | F1-score | IoU | F1-score | IoU | F1-score | IoU | F1-score | IoU | F1-score | mIoU | F1-score |
| Baseline (Deeplab-v3) [14] | 9.30 | 16.86 | 49.18 | 65.93 | 38.51 | 55.60 | 7.67 | 14.24 | 29.32 | 45.34 | 36.96 | 53.97 | 28.49 | 41.99 |
| BiSeNet only [11] | **29.01** | **44.97** | 22.70 | 36.99 | 0.69 | 1.36 | **41.56** | **58.71** | 26.12 | 41.42 | 20.93 | 34.61 | 23.50 | 36.34 |
| SEANet [39] | 13.16 | 23.23 | 51.33 | 67.79 | 42.44 | 59.56 | 5.15 | 9.79 | 29.14 | 45.13 | 39.61 | 56.73 | 30.14 | 43.70 |
| AdaptSegNet [35] | 8.36 | 15.33 | 49.55 | 64.64 | 40.95 | 58.11 | 22.59 | 36.79 | 34.43 | 61.50 | 48.01 | 63.41 | 33.98 | 49.96 |
| GAN-RSDA [50] | 27.39 | 43.43 | 18.66 | 20.71 | 0.59 | 1.82 | 32.06 | 31.93 | 19.72 | 31.08 | 27.40 | 24.24 | 20.97 | 25.54 |
| ProDA [37] | 10.63 | 19.21 | 44.70 | 61.72 | 46.78 | 63.74 | 31.59 | 48.02 | 40.55 | 57.71 | 56.85 | 72.49 | 38.51 | 53.82 |
| DualGAN [49] | 11.48 | 20.56 | 51.01 | 67.53 | 48.49 | 65.31 | 34.98 | 51.82 | 36.50 | 53.48 | 53.37 | 69.59 | 39.30 | 54.71 |
| Adaptation from Labeled Source Patches (LSPs) to Easy-to-adapt Target Patches (ETPs) | | | | | | | | | | | | | | |
| Baseline with $D_F$ (ours) | 13.99 | 24.55 | 55.58 | 71.45 | 44.54 | 61.63 | 30.73 | 47.02 | 40.67 | 57.82 | 54.93 | 70.91 | 40.07 | 55.56 |
| SL-Adapted-Baseline with $D_E$ (ours) | 12.34 | 21.97 | 60.76 | 75.59 | 53.94 | 70.08 | 31.80 | 48.26 | 43.62 | 60.74 | 61.37 | 76.06 | 43.97 | 58.78 |
| Adaptation from LSPs and Pseudo-labeled ETPs to Hard Target Patches | | | | | | | | | | | | | | |
| Baseline with $D_F$ (ours) | 11.77 | 21.06 | 63.57 | 77.73 | 59.20 | 74.37 | 32.84 | 49.44 | 46.38 | **63.37** | 67.35 | 80.49 | 46.85 | 61.08 |
| SL-Adapted-Baseline with $D_E$ (ours) | 12.31 | 24.59 | **64.39** | **78.59** | 59.35 | 75.08 | 37.55 | 54.60 | **47.17** | 63.27 | 66.44 | 79.84 | **47.87** | **62.66** |

For the existing domain adaptive semantic segmentation methods including BiSeNet [11], SEANet [39], AdaptSegNet [35], and GAN-RSDA [50], we report their experimental results which are implemented by DualGAN [49]. For ProDA [37] and DualGAN [49], we follow the hyper-parameter settings in their original paper, and the input size of each image, learning rate, momentum, weight decay, and batch size are set to $512 \times 512$, 0.0005, 0.9, 0.0005, and 4, respectively.

*2) Evaluation Metric:* For quantitative evaluation, since the intersection-over-union (IoU) can comprehensively evaluate the segmentation model including the segmentation results in terms of recall rate and precision rate [11]-[16], we use the IoU metric of each category and the mean value of the IoU (mIoU) to evaluate all methods. The IoU can be calculated as follows:

$$IoU = \frac{\hat{Y}_t \cap Y_t}{\hat{Y}_t \cup Y_t}. \quad (9)$$

where $\hat{Y}_t$ denotes the pixel-level prediction results obtained by the proposed method, and $Y_t$ denotes the pixel-level ground truth annotation from the target domain, which can only be used for model evaluation. Besides, following the setting in [49], we use F1-score to evaluate the proposed method, and the best model is selected according to its segmentation accuracy on the validation set from the target domain, and we report the results on the test set from the target domain.

*D. Experimental Results*

Considering that RSIs have their special characteristics such as the imaging mode differences, color saturation variations, and geographical location changes, we conduct the experiments in two cross-domain scenarios: 1) only imaging modes are dynamically changing; 2) both geographic locations and imaging modes are dynamically changing.

*1) Adaptation for Different Imaging Modes:* For this cross-domain adaptation scenario, Potsdam IR-R-G and Vaihingen IR-R-G are first employed as the source domain and target domain, respectively. The improvement in the segmentation accuracy in Table I is mainly due to two main reasons. For the first item, by comparing Baseline and Baseline with $D_F$, it can be seen from Table I that matching local semantic feature differences between domains can significantly boost the mIoU on the target domain. Next, the segmentation accuracy can be further improved when the global long-range feature discrepancies are reduced by comparing the Baseline with $D_F$ and SL-Adapted-Baseline with $D_E$. For the second item, an adaptation strategy that considers the uncertainty of patches from the target domain is beneficial to alleviate the domain discrepancies in this cross-domain scenario. It also can be observed from Table I that the cross-domain segmentation accuracy can be further increased by 4.48% mIoU when the model is adapted to hard-to-adapt target patches.

TABLE III
ADAPTATION FROM POTSDAM R-G-B TO VAIHINGEN IR-R-G. THE DEFINITION OF $D_F$, $D_E$ FOLLOWS TABLE I

| Method | Clutter/background | | Impervious surfaces | | Car | | Tree | | Low vegetation | | Building | | Overall | |
|---|---|---|---|---|---|---|---|---|---|---|---|---|---|---|
| | IoU | F1-Score | IoU | F1-score | IoU | F1-score | IoU | F1-score | IoU | F1-score | IoU | F1-score | mIoU | F1-score |
| Baseline (Deeplab-v3) [14] | 1.03 | 2.03 | 46.39 | 63.37 | 27.33 | 42.93 | 13.58 | 23.73 | 4.61 | 8.82 | 49.39 | 66.12 | 23.72 | 34.50 |
| BiSeNet only [11] | 0.88 | 1.75 | 36.58 | 53.57 | 4.81 | 9.17 | 7.74 | 14.37 | 18.73 | 31.55 | 17.30 | 29.49 | 14.34 | 23.30 |
| SEANet [39] | 7.05 | 12.92 | 41.84 | 59.00 | 31.72 | 48.16 | 21.86 | 35.82 | 20.83 | 34.48 | 56.80 | 72.44 | 30.02 | 43.80 |
| AdaptSegNet [35] | 2.99 | 5.81 | 51.26 | 67.77 | 10.25 | 18.54 | 51.51 | 68.02 | 12.75 | 22.61 | 60.72 | 75.55 | 31.58 | 43.05 |
| GAN-RSDA [50] | 2.27 | 4.43 | 34.55 | 51.35 | 9.90 | 18.00 | 24.94 | 39.89 | 22.58 | 36.83 | 40.64 | 57.79 | 22.48 | 34.70 |
| ProDA [37] | 2.39 | 5.09 | 49.04 | 66.11 | 31.56 | 48.16 | 49.11 | 65.86 | 32.44 | 49.06 | 68.94 | 81.89 | 38.91 | 52.70 |
| DualGAN [49] | 3.94 | 13.88 | 46.19 | 61.33 | 40.31 | 57.88 | **55.82** | **70.66** | 27.85 | 42.17 | 65.44 | 83.00 | 39.93 | 54.82 |
| Adaptation from Labeled Source Patches (LSPs) to Easy-to-adapt Target Patches (ETPs) | | | | | | | | | | | | | | |
| Baseline with $D_F$ (ours) | 1.79 | 4.13 | 60.22 | 74.33 | 41.20 | 58.05 | 48.97 | 66.29 | 27.58 | 42.63 | 69.75 | 81.64 | 41.58 | 54.51 |
| SL-Adapted-Baseline with $D_E$ (ours) | 4.04 | 9.14 | 63.41 | 77.01 | 38.71 | 55.40 | 53.08 | 69.99 | 37.86 | 54.75 | 76.16 | 86.40 | 45.54 | 58.78 |
| Adaptation from LSPs and Pseudo-labeled ETPs to Hard Target Patches | | | | | | | | | | | | | | |
| Baseline with $D_F$ (ours) | 9.88 | 18.64 | 62.84 | 76.89 | 40.24 | 57.15 | 50.88 | 68.10 | **39.03** | **56.26** | 73.88 | 84.81 | 46.13 | 60.31 |
| SL-Adapted-Baseline with $D_E$ (ours) | **12.38** | **21.55** | **64.47** | **77.76** | **43.43** | **60.05** | 52.83 | 69.62 | 38.37 | 55.94 | **76.87** | **86.95** | **48.06** | **61.98** |

TABLE IV
ADAPTATION FROM VAIHINGEN IR-R-G TO POTSDAM R-G-B. THE DEFINITION OF $D_F$, $D_E$ FOLLOWS TABLE I

| Method | Clutter/background | | Impervious surfaces | | Car | | Tree | | Low vegetation | | Building | | Overall | |
|---|---|---|---|---|---|---|---|---|---|---|---|---|---|---|
| | IoU | F1-Score | IoU | F1-score | IoU | F1-score | IoU | F1-score | IoU | F1-score | IoU | F1-score | mIoU | F1-score |
| Baseline (Deeplab-v3) [14] | 6.99 | 13.04 | 42.98 | 60.12 | 38.01 | 55.08 | 0.53 | 1.06 | 1.59 | 3.13 | 29.09 | 45.05 | 19.86 | 29.58 |
| BiSeNet only [11] | 23.66 | 38.26 | 17.74 | 30.12 | 0.99 | 1.95 | 32.67 | 49.24 | 18.42 | 31.11 | 12.64 | 22.43 | 17.69 | 28.85 |
| SEANet [39] | 4.68 | 8.75 | 28.90 | 44.77 | 44.77 | 61.52 | 5.17 | 9.82 | 8.70 | 15.88 | 36.24 | 53.20 | 21.36 | 32.32 |
| AdaptSegNet [35] | 6.11 | 11.50 | 37.66 | 59.55 | 42.31 | 55.95 | 30.71 | 45.41 | 15.10 | 25.81 | 54.25 | 70.31 | 31.02 | 44.75 |
| GAN-RSDA [50] | **27.75** | **43.43** | 11.56 | 20.71 | 0.92 | 1.82 | 19.00 | 31.93 | 18.40 | 31.08 | 13.79 | 24.24 | 15.24 | 25.54 |
| ProDA [37] | 11.13 | 20.51 | 44.77 | 62.03 | 41.21 | 59.27 | 30.56 | 46.91 | 35.84 | 52.75 | 46.37 | 63.06 | 34.98 | 50.76 |
| DualGAN [49] | 13.56 | 23.84 | 45.96 | 62.97 | 39.71 | 56.84 | 25.80 | 40.97 | **41.73** | **58.87** | 59.01 | 74.22 | 37.63 | 52.95 |
| Adaptation from Labeled Source Patches (LSPs) to Easy-to-adapt Target Patches (ETPs) | | | | | | | | | | | | | | |
| Baseline with $D_F$ (ours) | 4.68 | 8.94 | 48.35 | 65.18 | 38.90 | 56.01 | 36.89 | 53.89 | 27.20 | 42.76 | 53.04 | 69.31 | 34.84 | 49.35 |
| SL-Adapted-Baseline with $D_E$ (ours) | 8.27 | 15.27 | 56.06 | 71.84 | 50.79 | 67.37 | 35.64 | 52.55 | 24.69 | 39.61 | 63.13 | 77.40 | 39.76 | 54.01 |
| Adaptation from LSPs and Pseudo-labeled ETPs to Hard Target Patches | | | | | | | | | | | | | | |
| Baseline with $D_F$ (ours) | 10.39 | 19.48 | 56.68 | 72.35 | 55.63 | 71.49 | 35.41 | 52.30 | 25.20 | 40.26 | 63.36 | 77.57 | 41.11 | 55.58 |
| SL-Adapted-Baseline with $D_E$ (ours) | 13.27 | 23.43 | **57.65** | **73.14** | **56.99** | **72.27** | 35.87 | 52.80 | 29.77 | 45.88 | **65.44** | **79.11** | **43.17** | **57.77** |

Furthermore, we compare the proposed method optimized by using the CCDA strategy with state-of-the-art domain adaptive semantic segmentation methods including GAN-RSDA [50], DualGAN [49], BiSeNet [11], SEANet [39], AdaptSegNet [35], and ProDA [37] in this cross-domain scenario, where the first two are the state-of-the-art remote-sensing-images-based domain adaptive semantic segmentation methods and the last four represent natural-images-based UDA methods. Besides, GAN-RSDA [50] and DualGAN [49] design a GAN-based feature transferability network and an image-level style transfer network optimized via dynamically adjusting the constraint weights, respectively. BiSeNet [11] proposes a bilateral segmentation path and SEANet [39] exploits a self-ensembling attention network to inherit the advantages in both ensemble model and self-training, and can reduce the domain gaps and improve the feature transferability of semantic segmentation network. AdaptSegNet [35] aligns the distribution of features in the output space, representing a method that reduces the inter-domain feature discrepancies from globally spatial relationships. ProDA [37] tackle the cross-domain adaptation problem from a new perspective of self-training. For a fair comparison, the experimental settings of all methods are consistent, and we compare the SL-Adapted-Baseline with $D_E$, namely, the baseline using the CCDA strategy and LGFA module, with DualGAN, SEANet, AdaptSegNet, ProDA. The results indicate that, compared with the state-of-the-art cross-domain methods that fail to progressively adapt the model according to the uncertainty of target patches, the segmentation accuracy gains from the CCDA strategy and LGFA module are considerable.

TABLE V
ADAPTATION RESULTS FROM POTSDAM IR-R-G TO VAIHINGEN IR-R-G FOR ANOTHER BASELINE SEGMENTATION NETWORK. THE DEFINITION OF $D_F$, $D_E$ FOLLOWS TABLE I

| Method | Impervious surfaces | | Car | | Tree | | Low vegetation | | Building | | Overall | |
|---|---|---|---|---|---|---|---|---|---|---|---|---|
| | IoU | F1-Score | IoU | F1-score | IoU | F1-score | IoU | F1-score | IoU | F1-score | IoU | F1-score |
| Baseline (UNet) [15] | 12.87 | \ | 3.69 | \ | 10.98 | \ | 9.20 | \ | 28.12 | \ | 12.97 | 25.95 |
| AdaptSegNet [35] | 32.63 | \ | 11.13 | \ | **54.27** | \ | **35.29** | \ | 37.38 | \ | 34.14 | 54.66 |
| BSANet (UNet) [45] | 53.55 | \ | 27.24 | \ | 51.93 | \ | 23.91 | \ | 59.87 | \ | 43.30 | 62.68 |
| Adaptation from Labeled Source Patches (LSPs) to Easy-to-adapt Target Patches (ETPs) | | | | | | | | | | | | |
| Baseline with $D_F$ (ours) | 52.36 | 68.73 | 19.00 | 31.93 | 48.80 | 65.59 | 27.83 | 43.55 | 53.11 | 69.37 | 40.22 | 55.83 |
| SL-Adapted-Baseline with $D_E$ (ours) | 56.18 | 71.95 | 22.94 | 37.32 | 51.33 | 67.83 | 25.46 | 40.59 | 55.19 | 71.63 | 42.22 | 57.86 |
| Adaptation from LSPs and Pseudo-labeled ETPs to Hard Target Patches | | | | | | | | | | | | |
| Baseline with $D_F$ (ours) | 57.47 | 72.99 | **28.61** | 44.78 | 53.29 | **69.53** | 24.83 | 39.78 | 59.81 | 75.27 | 44.80 | 60.47 |
| SL-Adapted-Baseline with $D_E$ (ours) | **58.64** | **75.13** | 28.17 | **45.81** | 53.28 | 69.52 | 30.39 | **47.62** | **60.60** | **76.89** | **46.22** | **62.99** |

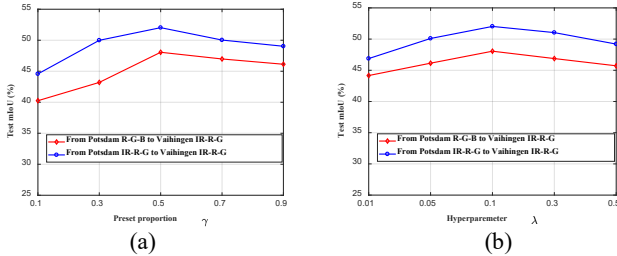

Fig. 4. Ablation study of each design choice including sensitivity analyses for (a) preset proportion $\gamma$, and (b) hyper-parameter $\lambda$.

TABLE VI
THE STUDY OF ADAPTATION ORDER WHEN $\gamma$ =0.5 AND $\lambda$ =0.1, WHERE POTSRGB AND VAIH REPRESENT THE POTSDAM R-G-B AND VAIHINGEN IR-R-G DATASETS, RESPECTIVELY.

| Adaptation order of the target domain | PotsRGB to Vaih | PotsIRRG to Vaih |
|---|---|---|
| Without curriculum-style learning | 40.66 | 43.00 |
| Reverse the order of feature adaptation | 42.26 | 45.91 |
| Reverse the order of patch adaptation | 45.45 | 46.60 |
| Reverse both feature and patch adaptations | 37.96 | 41.13 |
| With curriculum-style learning | **48.06** | **52.03** |

On the other hand, we change the cross-domain setting where Vaihingen IR-R-G and Potsdam IR-R-G are employed as the source domain and target domain, respectively, and repeat the above experiments (the results are shown in Table II). It can be concluded from Table II that the segmentation mIoU achieves consistent improvements, showing the generalization ability of the proposed method on the different cross-domain settings.

*2) Adaptation for Different Geographic Locations and Imaging Modes:* To validate that the proposed method can be generalized to different cross-domain adaptation scenarios, we conduct experiments in another common scenario. Tables III and IV report the results of adaptation from Potsdam R-G-B to Vaihingen IR-R-G and from Vaihingen IR-R-G to Potsdam R-G-B, respectively. It also can be seen that the proposed method strengthens the feature transferability of the baseline segmentation network for the above datasets. Next, the proposed method is compared with cross-domain methods including GAN-RSDA [50], DualGAN [49], SEANet [39], AdaptSegNet [35], and ProDA [37]. The experimental results also demonstrate that the proposed method exceeds all existing cross-domain methods with a considerable margin, further validating the effectiveness of our method for RSIs-based semantic segmentation.

IV. INSIGHT ANALYSES AND DISCUSSION

*A. Ablation Study*

In order to analyze the effectiveness of each design choice, we conduct the ablation studies from two aspects: First, the transferability of semantic segmentation network when changing the preset proportion of easy-to-adapt target patches in the total target patches $\gamma$; Second, the impact of changing the hyper-parameter $\lambda$ in Eqs. (7) and (8) on segmentation accuracy. Experimental results in Figs. 4(a) and 4(b) show that our method achieves the best mIoU when $\gamma$ and $\lambda$ equal to 0.5 and 0.1, respectively.

*B. Generalization Evaluation for CCDA Strategy*

To comprehensively validate the generalization ability of the proposed CCDA strategy using the LGFA module on different segmentation baselines, we conduct experiments by applying the proposed method to UNet [15], which can be regarded as a representative model in the field of medical image segmentation task. Specifically, following the experimental setup introduced in Section III.C, we first train UNet on the annotation-rich source data. Next, the UNet well-trained on its source domain is progressively adapted to unlabeled target patches (easy-to-adapt and hard-to-adapt target patches during the first and second stages, respectively) according to the CCDA strategy, and the feature alignment process can be performed via the LGFA module to alleviate the inter-domain discrepancy of UNet.

Recently, BASNet [43] also employs the UNet as baseline segmentation network, and designs a dual-branch network structure to align the cross-domain feature discrepancies from the image domain and the wavelet domain simultaneously. To fairly compare with BASNet [43], we also ignore the clutter category and only report the experimental results of the remaining five categories on Vaihingen dataset. The results in Table V show that the proposed CCDA strategy and LGFA module are also beneficial to a different segmentation baseline model, further validating its generalization ability.

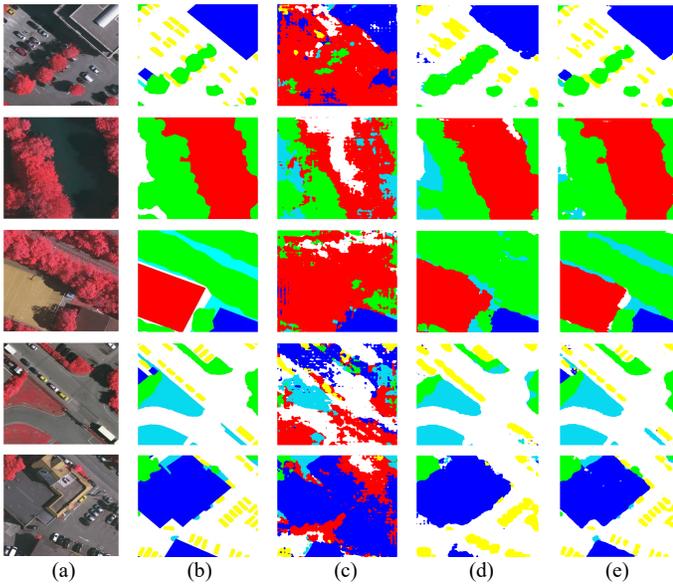

Fig. 5. Visual segmentation results of easy-to-adapt patches on Vaihingen IR-R-G. (a) Input images. (b) Ground truth. (c) Results without the domain adaptation. (d) Results predicted by the first stage of the proposed method (adapt to easy target patches). (e) Results predicted by the proposed method using all developed modules.

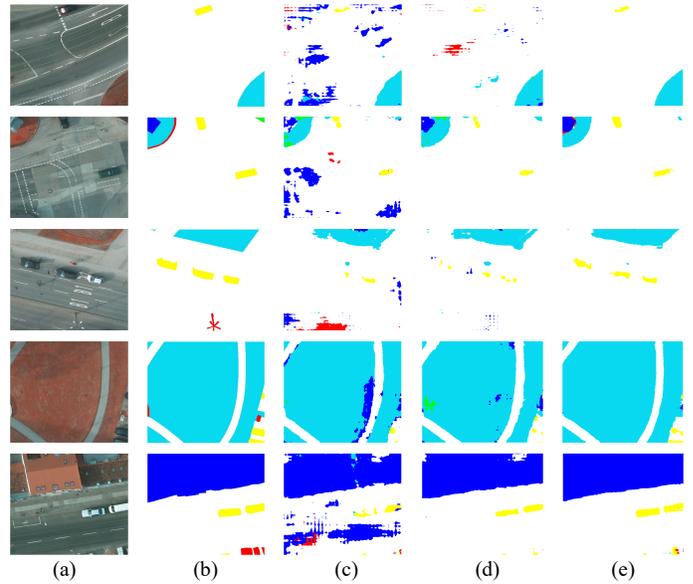

Fig. 7. Visual segmentation results of easy-to-adapt patches on Potsdam IR-R-G. (a) Input images. (b) Ground truth. (c) Results without the domain adaptation. (d) Results predicted by the first stage of the proposed method (adapt to easy target patches). (e) Results predicted by the proposed method using all developed modules.

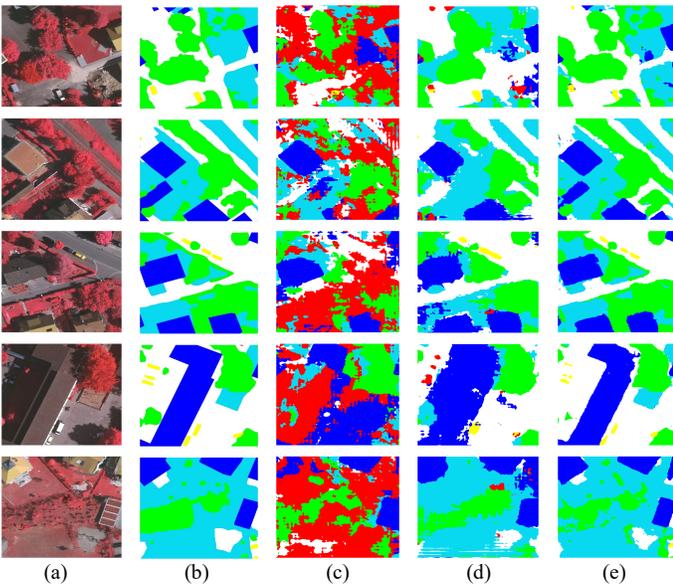

Fig. 6. Visual segmentation results of hard-to-adapt patches on Vaihingen IR-R-G. (a) Input images. (b) Ground truth. (c) Results without the domain adaptation. (d) Results predicted by the first stage of the proposed method (adapt to easy target patches). (e) Results predicted by the proposed method using all developed modules.

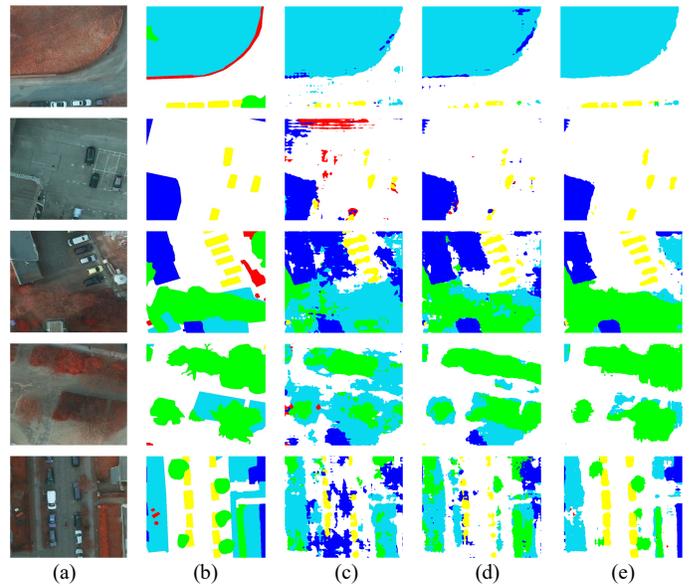

Fig. 8. Visual segmentation results of hard-to-adapt patches on Potsdam IR-R-G. (a) Input images. (b) Ground truth. (c) Results without the domain adaptation. (d) Results predicted by the first stage of the proposed method (adapt to easy target patches). (e) Results predicted by the proposed method using all developed modules.

### C. The Role of Curriculum-style Adaptation

In order to further investigate the impact of sample or feature adaptation order (learning order) on model cross-domain adaptation, we manually change the order of sample or feature adaptation from the following aspects: 1) Reverse the order of feature-level adaptation, namely, the adaptation from *global* to *local* feature variations; 2) Reverse the order of patch-level adaptation, from *hard* to *easy* patches; 3) Reverse the order of both feature- and patch-level adaptations. It can be seen from Table VI that, reversing the adaptation order of both features and patches seriously hurts the model accuracy, and its results are even worse than those without a designed curriculum. This is mainly due to that a large number of incorrect pseudo-labels in the first stage bring prediction deviations for data distribution from the target domain, further resulting in a more severe *negative transfer* occurred in the second stage, indicating that a well-designed adaptation curriculum is necessary for the VHR RSIs-based cross-domain adaptation. Moreover, by means of an effective adaptation order for both target features and patches, the best segmentation accuracy of 48.06% and 52.03% can be achieved.

*D. Visual Image Segmentation Results*

We show visual examples of different segmentation models on both the easy-to-adapt patches and the hard-to-adapt patches of Vaihingen IR-R-G (shown in Figs. 5 and 6) and Potsdam IR-R-G (shown in Figs. 7 and 8). These visualization results show a great superiority of our method under different baseline methods. Furthermore, it can be seen from Fig. 5 to Fig. 8 that the first-stage adaptation, which aims at adapting the model to easy target patches and further is helpful to reduce the risk of negative transfer, can greatly boost the segmentation visualization results on the target domain compared with the methods without domain adaptation. After the end of the adaptation process in the first stage, the model can be further transferred to hard-to-adapt target domain.

## V. CONCLUSIONS

In this paper, we aim to transfer a pre-trained RSIs-based segmentation model from its source domain to an unlabeled target domain, and have proposed a curriculum-style cross-domain adaptation (CCDA) strategy and a local-to-global feature alignment (LGFA) module. Considering that the patch-wise data distribution of remote sensing images often presents serious intra-domain feature differences, the CCDA strategy is developed to alleviate such domain discrepancies within a remote sensing image from the target domain, by means of the progressive adaptation way using an entropy-based score assigned to each target patch. Furthermore, due to that the locally semantic features and globally spatial layouts are dynamically changing in RSIs, the LGFA module is adopted to reduce such feature differences from local details and spatial layouts by means of well-designed semantic-level and entropy-level domain classifiers, respectively. The experimental results on common benchmarks have demonstrated the effectiveness and superiority of the proposed method in RSIs-based cross-domain scenarios compared with the state-of-the-art methods.

With the launch of satellites carrying higher resolution sensors, large amounts of unlabeled data from the target domain will be available. In this case, few-shot adaptation using only a few manually annotated samples from the new target domain can improve the cross-domain transferability of features under a dynamically changing imaging sensor, which will be a possible research topic in our future work.